\documentclass[10pt, conference, compsocconf]{IEEEtran}
\pdfoutput=1 
\usepackage{blindtext, graphicx}
\usepackage{epsfig}
\usepackage{syntonly}
\usepackage{rotating}
\usepackage{amsmath}
\usepackage{setspace}
\usepackage{verbatim} 
\usepackage{amssymb}
\usepackage{amsmath}
\usepackage{graphicx}
\usepackage{multirow}
\usepackage{colortbl}
\usepackage{enumerate}
\usepackage{color}
\usepackage{wrapfig}

\usepackage{algorithm2e}
\usepackage{cleveref}
\usepackage{comment}
\usepackage{ifthen}
\usepackage[dotinlabels]{titletoc}
\usepackage{colortbl}
\usepackage{array}
\usepackage{alltt}
\usepackage{bm}

\usepackage[normalem]{ulem}

\pdfoutput=1 
\usepackage{blindtext, graphicx}
\usepackage{epsfig}
\usepackage{syntonly}
\usepackage{rotating}
\usepackage{amsmath}
\usepackage{setspace}
\usepackage{verbatim} 
\usepackage{amssymb}
\usepackage{amsmath}
\usepackage{graphicx}
\usepackage{multirow}
\usepackage{colortbl}
\usepackage{enumerate}
\usepackage{color}
\usepackage{wrapfig}
\usepackage{algorithm2e}
\usepackage{cleveref}
\usepackage{comment}
\usepackage{ifthen}
\usepackage[dotinlabels]{titletoc}
\usepackage{colortbl}
\usepackage{array}

\usepackage{alltt}
\usepackage{bm}
\usepackage{longtable,booktabs}
\usepackage{url}
\usepackage{courier}
\usepackage{cite}

\newcommand{\authemail}[1]{\normalfont{\texttt{#1}}}

\newcommand{\Affil}[1]{\fontsize{11}{11}\itshape{#1}}
\usepackage{tabularx}
\usepackage{footnote}

\begin{document}
\title{EIE: Efficient Inference Engine on Compressed Deep Neural Network}

\author{
\IEEEauthorblockN{
	\large{Song Han\IEEEauthorrefmark{1}}
	Xingyu Liu\IEEEauthorrefmark{1}
	Huizi Mao\IEEEauthorrefmark{1} 
	Jing Pu\IEEEauthorrefmark{1}
	Ardavan Pedram\IEEEauthorrefmark{1}\\
	Mark A. Horowitz\IEEEauthorrefmark{1}
	William J. Dally\IEEEauthorrefmark{1}\IEEEauthorrefmark{2}
}
\IEEEauthorblockA{\Affil{\IEEEauthorrefmark{1}Stanford University}, 
\IEEEauthorrefmark{2}NVIDIA}
\authemail{\{songhan,xyl,huizi,jingpu,perdavan,horowitz,dally\}@stanford.edu}
}
\maketitle

\begin{abstract}

State-of-the-art deep neural networks (DNNs) have hundreds of millions of connections and are both computationally and memory intensive, making them difficult to deploy on embedded systems with limited hardware resources and power budgets.  While custom hardware helps the computation, fetching weights from DRAM is two orders of magnitude more expensive than ALU operations, and dominates the required power. 

Previously proposed `Deep Compression' makes it possible to fit large DNNs (AlexNet and VGGNet) fully in on-chip SRAM. This compression is achieved by pruning the redundant connections and having multiple connections share the same weight. We propose an energy efficient inference engine (EIE) that performs inference on this compressed network model and accelerates the resulting sparse matrix-vector multiplication with weight sharing. Going from DRAM to SRAM gives EIE 120$\times$ energy saving; Exploiting sparsity saves 10$\times$; Weight sharing gives 8$\times$; Skipping zero activations from ReLU saves another 3$\times$. Evaluated on nine DNN benchmarks, EIE is 189$\times$ and 13$\times$ faster when compared to CPU and GPU implementations of the same DNN without compression. EIE has a processing power of 102~GOPS/s working directly on a compressed network, corresponding to 3~TOPS/s on an uncompressed network, and processes FC layers of AlexNet at 1.88$\times$10\textsuperscript{4} frames/sec with a power dissipation of only 600mW.  It is 24,000$\times$ and 3,400$\times$ more energy efficient than a CPU and GPU respectively. Compared with DaDianNao, EIE has 2.9$\times$, 19$\times$ and 3$\times$ better throughput, energy efficiency and area efficiency. 
\end{abstract}


\IEEEpeerreviewmaketitle

\section{Introduction}

Neural networks have become ubiquitous in applications including computer vision~\cite{hinton12,szegedy2014GoogLenet,simonyan2014very}, speech recognition~\cite{mikolov2010RNN}, and natural language processing~\cite{mikolov2010RNN}. In 1998, Lecun et al. classified handwritten digits with less than 1M parameters~\cite{lecun1998gradient}, while in 2012, Krizhevsky et al. won the ImageNet competition with 60M parameters~\cite{hinton12}. Deepface classified human faces with 120M parameters~\cite{deepface}. Neural Talk~\cite{neural_talk} automatically converts image to natural language with 130M CNN parameters and 100M RNN parameters. Coates et al. scaled up a network to 10 billion parameters on HPC systems~\cite{cots}.

Large DNN models are very powerful but consume large amounts of energy because the model must be stored in external DRAM, and fetched every time for each image, word, or speech sample. For embedded mobile applications, these resource demands become prohibitive. Table \ref{energy} shows the energy cost of basic arithmetic and memory operations in a 45nm CMOS process~\cite{markenergy}. It shows that the total energy is dominated by the required memory access if there is no data reuse. The energy cost per fetch ranges from 5pJ for 32b coefficients in on-chip SRAM to 640pJ for 32b coefficients in off-chip LPDDR2 DRAM. Large networks do not fit in on-chip storage and hence require the more costly DRAM accesses. Running a 1G connection neural network, for example, at 20Hz would require $(20Hz)(1G)(640pJ) = 12.8W$ just for DRAM accesses, which is well beyond the power envelope of a typical mobile device.

\begin{figure}[t!]
\centering
\vspace{10pt}
\scalebox{1}[1.1]{\includegraphics[width=0.5\textwidth]{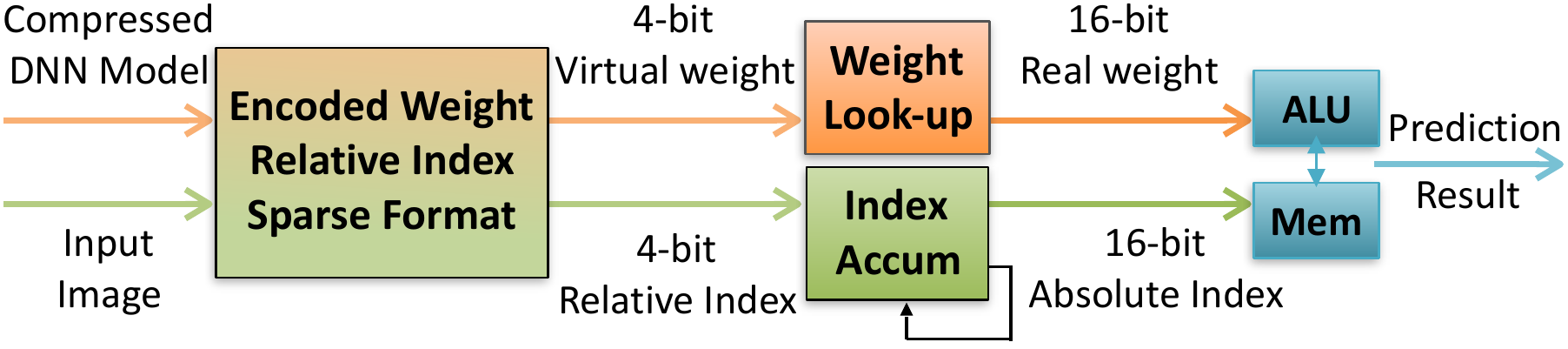}}
\caption{Efficient inference engine that works on the compressed deep neural network model for machine learning applications. }
\label{fig:begin}
\vspace{-15pt}
\end{figure}

Previous work has used specialized hardware to accelerate DNNs~\cite{diannao,dadiannao,shidiannao}. However, these efforts focus on accelerating dense, uncompressed models - limiting their utility to small models or to cases where the high energy cost of external DRAM access can be tolerated. Without model compression, it is only possible to fit very small neural networks, such as Lenet-5, in on-chip SRAM~\cite{shidiannao}.

Efficient implementation of convolutional layers in CNN 
has been intensively studied, as its data reuse and manipulation is quite suitable for customized hardware~\cite{diannao,dadiannao,shidiannao,farabet2009cnp,fpga2016cnn,isaac}.
However, it has been found that fully-connected (FC) layers, widely used in RNN and LSTMs, are bandwidth limited on large networks~\cite{fpga2016cnn}. Unlike CONV layers, there is no parameter reuse in FC layers. Data batching has become an efficient solution when training networks on CPUs or GPUs, however, it is unsuitable for real-time applications with latency requirements.

\begin{table} [!htbp]
\footnotesize
\caption{Energy table for 45nm CMOS process \cite{markenergy}. DRAM access uses three orders of magnitude more energy than simple arithmetic
and 128x more than SRAM.}
\vspace{-5pt}
\centering
\begin{tabular}{@{}lll@{}}
\toprule
Operation            & Energy {[}pJ{]} & Relative Cost \\ \midrule
32 bit int ADD       & 0.1             & 1             \\
32 bit float ADD        & 0.9             & 9           \\
32 bit int MULT      & 3.1             & 31          \\
32 bit float MULT       & 3.7             & 37          \\
32 bit 32KB SRAM     & 5               & 50          \\
\bf{32 bit DRAM }  & \bf{640}        & \bf{6400}   \\ \bottomrule
\end{tabular}
\vspace{-15pt}

\label{energy}
\end{table}

Network compression via pruning and weight sharing \cite{song_pruning} makes it possible to fit modern networks such as AlexNet  (60M parameters, 240MB), and VGG-16  (130M parameters, 520MB) in on-chip SRAM.
Processing these compressed models, however, is challenging. With pruning, the matrix becomes sparse and the indicies become relative. With weight sharing, we store only a short (4-bit) index for each weight. This adds extra levels of indirection that cause complexity and inefficiency on CPUs and GPUs.

To efficiently operate on compressed DNN models,
we propose EIE, an efficient inference engine, a specialized accelerator that  performs customized sparse matrix vector multiplication and handles weight sharing with no loss of efficiency.
EIE is a scalable array of processing elements (PEs) 
Every PE stores a partition of network in SRAM and performs the computations associated with that part. It takes advantage of dynamic input vector sparsity, static weight sparsity, relative indexing, weight sharing and extremely narrow weights~(4bits).

In our design, each PE holds 131K weights of the compressed model, corresponding to 1.2M weights of the original dense model, and is able to perform 800 million weight calculations per second.  
In 45nm CMOS technology, an EIE PE has an area of 0.638mm$^2$ and dissipates 9.16mW at 800MHz. 
The fully-connected layers of AlexNet fit into 64PEs that consume a total of 40.8mm$^2$ and operate at $1.88\times10^4$ frames/sec with a power dissipation of 590mW.
Compared with CPU~(Intel i7-5930k) GPU~(GeForce TITAN X) and mobile GPU~(Tegra K1), EIE achieves $189\times$, $13\times$ and $307\times$  acceleration, while saving $24,000\times$, $3,400\times$ and $2,700\times$ energy, respectively.

This paper makes the following contributions:

\begin{enumerate}
\item We present the first accelerator for sparse and weight sharing neural networks. Operating directly on compressed networks enables the large neural network models to fit in on-chip SRAM, which results in $120\times$ better energy savings compared to accessing from external DRAM.

\item EIE is the first accelerator that exploits the dynamic sparsity of activations to save computation.  EIE saves 65.16\% energy by avoiding weight references and arithmetic for the 70\% of activations that are zero in a typical deep learning applications. 

\item We describe a method of both distributed storage and distributed computation to parallelize a sparsified layer across multiple PEs, which achieves load balance and good scalability. 

\item We evaluate EIE on a wide range of deep learning models, including CNN for object detection and LSTM for natural language processing and image captioning. 
We also compare EIE to CPU, GPU, FPGA, and other ASIC accelerators.

\end{enumerate}

 In Section~\ref{sec:motivation} we describe the motivation of accelerating compressed networks. Section~\ref{sec:compressed} reviews the compression technique that EIE uses and how to parallelize the workload. The hardware architecture in each PE to perform inference is described in Section~\ref{sec:hardware}. In Section~\ref{sec:evaluation} we describe our evaluation methodology. Then we report our experimental results in Section~\ref{sec:experimental}, followed by discussions, comparison with related work and conclusions. 
\section{Motivation}
\label{sec:motivation}

Matrix-vector multiplication~(M$\times$V) is a basic building block in a wide range of neural networks and deep learning applications. In convolutional neural network~(CNN), fully connected layers are implemented with M$\times$V, and more than 96\% of the connections are in the FC layers~\cite{hinton12}. In object detection algorithms, an FC layer is required to run multiple times on all proposal regions, taking up to $38\%$ computation time~\cite{girshick2015fast}. In recurrent neural network~(RNN), M$\times$V operations are performed on the new input and the hidden state at each time step, producing a new hidden state and the output. Long-Short-Term-Memory~(LSTM)~\cite{lstm} is a widely used structure of RNN cell that provides more complex hidden unit computation. Each LSTM cell can be decomposed into eight M$\times$V operations, two for each: input gate, forget gate, output gate, and one temporary memory cell. RNN, including LSTM, is widely used in image captioning~\cite{neural_talk}, speech recognition~\cite{graves2005framewise} and natural language processing~\cite{mikolov2010RNN}. 

During M$\times$V, the memory access is usually the bottleneck~\cite{fpga2016cnn} especially when the matrix is larger than the cache capacity. There is no reuse of the input matrix, thus a memory access is needed for every operation. On CPUs and GPUs, this problem is usually solved by batching, i.e., combining multiple vectors into a matrix to reuse the parameters. However, such a strategy is unsuitable for real-time applications that are latency-sensitive, such as pedestrian detection in an autonomous vehicle\cite{lane2015can}, because batching substantially increases latency. Therefore it is preferable to create an efficient method of executing large neural networks without the latency cost of batching.

Because memory access is the bottleneck in large layers, compressing the neural network offers a solution. Though compression reduces the total number of operations, the irregular pattern caused by compression hinders the effective acceleration on CPUs and GPUs, as illustrated in Table~\ref{table:detailed_performance}. 

A compressed network is not efficient on previous accelerators either. Previous SPMV accelerators can only exploit the static weight sparsity.  They are unable to exploit dynamic activation sparsity\cite{SpMxV:FPGA:2014}. Previous DNN accelerators cannot exploit either form of sparsity and must expand the network to dense form before operation\cite{dadiannao}. Neither is able to exploit weight sharing. This motivates building a special engine that can operate on a compressed network.

\section{DNN Compression and Parallelization}
\label{sec:compressed}

\subsection{Computation}

A FC layer of a DNN performs the computation
\begin{equation}\label{eq:eqn1}
\begin{aligned}
b = {f}(Wa+v)
\end{aligned}
\end{equation}

\noindent
Where $a$ is the input activation vector, $b$ is the output activation vector, $v$ is the bias, $W$ is the weight matrix, and $f$ is the  non-linear function, typically the Rectified Linear Unit(ReLU)\cite{relu} in CNN and some RNN. Sometimes $v$ will be combined with $W$ by appending an additional one to vector $a$, therefore we neglect the bias in the following paragraphs.

For a typical FC layer like FC7 of VGG-16 or AlexNet, the activation vectors are 4K long, and the weight
matrix is 4K $\times$ 4K (16M weights).
Weights are represented as single-precision floating-point numbers so such a layer requires 64MB of storage. The output activations of Equation~(1) are computed element-wise as:

\vspace{-5pt}
\begin{align}
b_i = { ReLU}\left(\sum_{j=0}^{n-1}W_{ij}a_j\right)\label{eqn2}
\end{align}
\vspace{-5pt}


{\em Deep Compression}~\cite{han2015deep} describes a method to compress DNNs without loss of accuracy through a combination of pruning and 
weight sharing.
Pruning makes matrix $W$ sparse with density $D$ ranging from 4\% to
25\% for our benchmark layers.
Weight sharing replaces each weight $W_{ij}$ with a four-bit index
$I_{ij}$ into a shared table $S$ of 16 possible weight values.

With deep compression, the per-activation computation of Equation~(2)
becomes

\vspace{-5pt}
\begin{align}
b_i = {ReLU}\left(\sum_{j \in X_i \cap Y}S[I_{ij}]a_j \right) \label{eqn3}
\end{align}
\vspace{-5pt}

\noindent
Where $X_i$ is the set of columns $j$ for which $W_{ij} \neq 0$,
$Y$ is the set of indices $j$ for which $a_j \neq 0$, $I_{ij}$
is the index to the shared weight that replaces $W_{ij}$, and $S$
is the table of shared weights.
Here $X_i$ represents the static sparsity of $W$ and $Y$ represents
the dynamic sparsity of $a$.
The set $X_i$ is fixed for a given model.
The set $Y$ varies from input to input.

Accelerating Equation~(3) is needed to accelerate
a compressed DNN. 
We perform the indexing $S[I_{ij}]$ and the multiply-add
only for those columns for which both $W_{ij}$ and $a_j$ are 
non-zero, so that both the sparsity of the matrix and the vector are exploited.
This results in a dynamically irregular computation.
Performing the indexing itself involves bit manipulations
to extract four-bit $I_{ij}$ and an extra load (which is 
almost assured a cache hit).

\subsection{Representation}\label{sec:CSC}

To exploit the sparsity of activations we store
our encoded sparse weight matrix $W$ in 
a variation of compressed sparse column (CSC)
format ~\cite{vuduc2003automatic}.

For each column $W_j$ of matrix $W$ we store a vector $v$ that
contains the non-zero weights, and a second, equal-length vector $z$ that
encodes the number of zeros before the corresponding entry in $v$.
Each entry of $v$ and $z$ is represented by a four-bit value.
If more than 15 zeros appear before a non-zero entry we add a zero
in vector $v$.
For example, we encode the following column

\vspace{-5pt}
$$
[0, 0, 1, 2, 0, 0, 0, 0, 0, 0, 0, 0, 0, 0, 0, 0, 0, 0, 0, \bm{0}, 0, 0, 3]
$$
\vspace{-5pt}

\noindent
as $v = [1, 2, \bm{0}, 3]$, $z = [2, 0, \bm{15}, 2]$.
$v$ and $z$ of all columns are stored in one large
pair of arrays with a pointer vector $p$ pointing to the 
beginning of the vector for each column.
A final entry in $p$ points one beyond the last vector element
so that
the number of non-zeros in column $j$ (including padded zeros) 
is given by $p_{j+1} - p_j$.

Storing the sparse matrix by columns in CSC format makes it easy to exploit activation sparsity.  
We simply multiply each non-zero activation by all of the non-zero
elements in its corresponding column.

\begin{figure}[!tb]
\centering
	\includegraphics[width=0.48\textwidth]{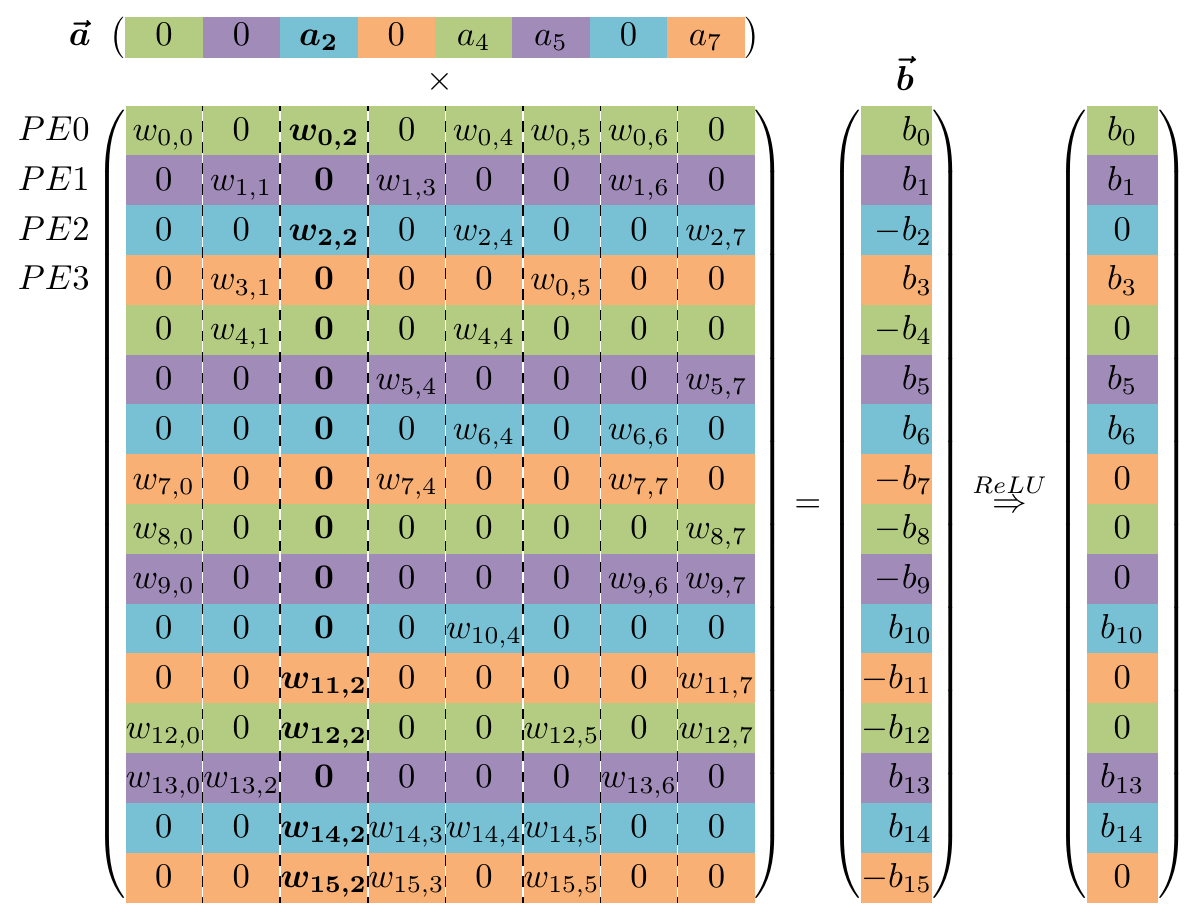}
      \vspace{-15pt}
	\caption{Matrix $W$ and vectors $a$ and $b$ are interleaved over 4 PEs. Elements of the same color are stored in the same PE.}
    	\vspace{-5pt}
	\label{fig:matrix}
\end{figure}

\begin{figure}[!tb]    
	\centering
	\vspace{-0pt}
    \includegraphics[width=0.5\textwidth]{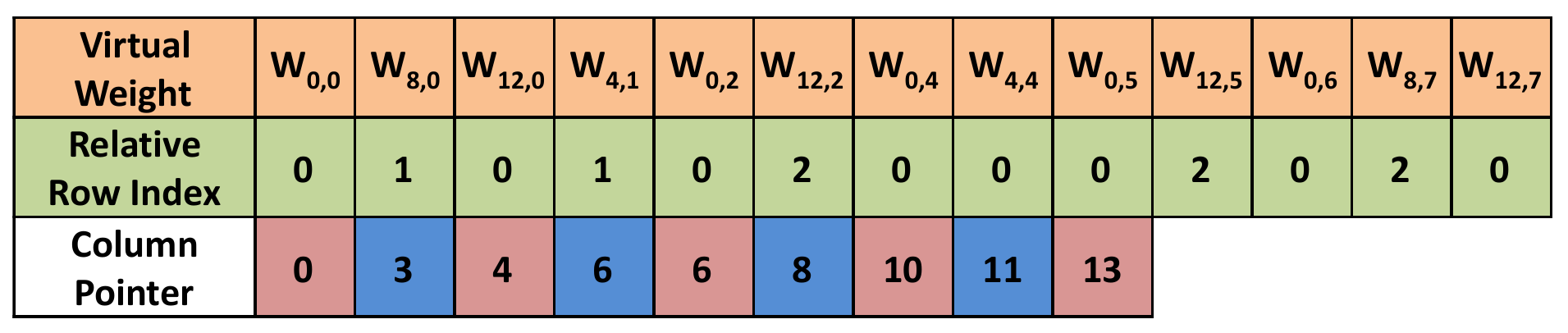}
        \vspace{-15pt}
	\caption{Memory layout for the relative indexed, indirect weighted and interleaved CSC format, corresponding to PE$_0$ in Figure~\ref{fig:matrix}. }
            \vspace{-10pt}

	\label{fig:example}
\end{figure}

\begin{figure*}[!tb]
    \vspace{-15pt}
	\centering
\includegraphics[width=1\textwidth]{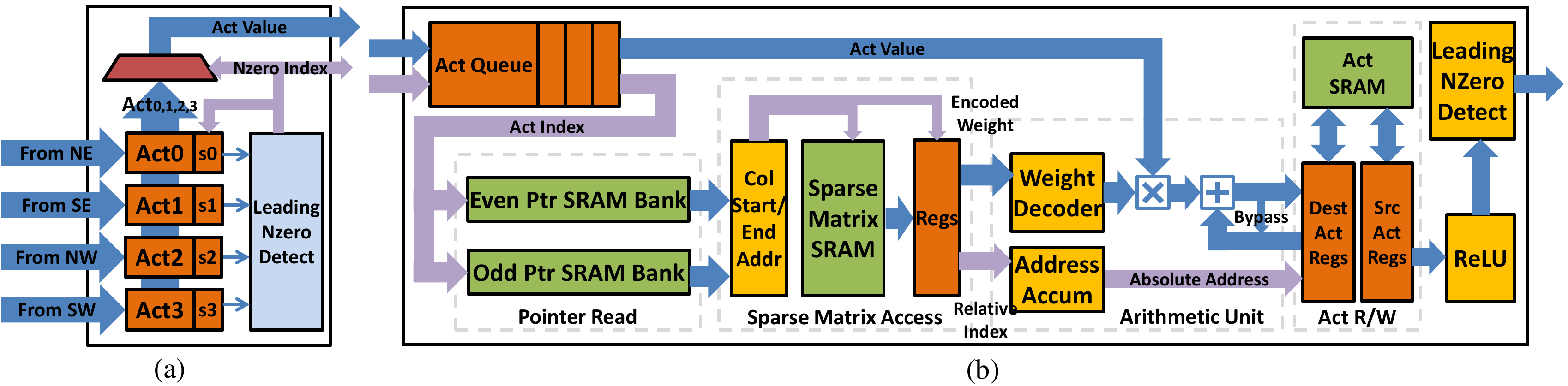} 			
	\vspace{-20pt}
\caption{(a) The architecture of  Leading Non-zero Detection Node. (b) The architecture of Processing Element.}
\label{fig:architectures}
 \vspace{-10pt}
\end{figure*}

\subsection{Parallelizing Compressed DNN}
\label{sec:parallelizing}

We distribute the matrix and parallelize our matrix-vector computation
by interleaving the rows of the matrix $W$ over multiple processing
elements (PEs).
With $N$ PEs, PE$_k$ holds all rows $W_i$, output activations
$b_i$, and input activations $a_i$ for which $i \pmod{N} = k$.
The portion of column $W_j$ in PE$_k$ is stored in the CSC format
described in Section~\ref{sec:CSC} but with the zero counts referring only to zeros in the subset of the column in this PE.
Each PE has its own $v$, $x$, and $p$ arrays that encode its
fraction of the sparse matrix.

Figure~\ref{fig:matrix} shows an example multiplying an input
activation vector $a$ (of length 8) by a $16\times8$ weight matrix $W$ 
yielding an output activation vector $b$ (of length 16) on $N=4$ PEs.  The elements of $a$, $b$, and $W$ are color coded with their PE assignments. Each PE owns 4 rows of $W$, 2 elements of $a$, and 4 elements of $b$. 

We perform the sparse matrix $\times$ sparse vector operation by scanning 
vector $a$ to find its next non-zero value $a_j$ and broadcasting $a_j$ along with its index $j$ to all PEs.
Each PE then multiplies $a_j$ by the non-zero elements in its portion of
column $W_j$ --- accumulating the partial sums in accumulators for
each element of the output activation vector $b$.
In the CSC representation these non-zeros weights are stored contiguously
so each PE simply walks through its $v$ array from location $p_j$ to 
$p_{j+1}-1$ to load the weights.
To address the output accumulators, the row number $i$ corresponding
to each weight $W_{ij}$ is generated by keeping a running sum
of the entries of the $x$ array.

In the example of Figure~\ref{fig:matrix}, the first non-zero is $a_2$ on $PE_2$.
The value $a_2$ and its column index $2$ is broadcast to all PEs.
Each PE then multiplies $a_2$ by every non-zero in its portion of column 2.
$PE_0$ multiplies $a_2$ by $W_{0,2}$ and $W_{12,2}$;
$PE_1$ has all zeros in column 2 and so performs no multiplications;
$PE_2$ multiplies $a_2$ by $W_{2,2}$ and $W_{14,2}$, and so on.
The result of each product is summed into the corresponding row accumulator.
For example $PE_0$ computes $b_0 = b_0 + W_{0,2}a_2$
and $b_{12} = b_{12}+ W_{12,2}a_2$.
The accumulators are initialized to zero before each layer computation.

The interleaved CSC representation facilitates exploitation of both the dynamic sparsity of activation vector $a$ and the static sparsity of the weight matrix $W$.
We exploit activation sparsity by broadcasting only non-zero elements of  input activation $a$.
Columns corresponding to zeros in $a$ are completely skipped.
The interleaved CSC representation allows each PE to quickly find
the non-zeros in each column to be multiplied by $a_j$.
This organization also keeps all of the computation except for the 
broadcast of the input activations local to a PE. The interleaved CSC representation of matrix in Figure \ref{fig:matrix} is shown in Figure \ref{fig:example}.

This process may suffer load imbalance because
each PE may have a different number of non-zeros in a particular column.
We will see in Section~\ref{sec:hardware} how this load imbalance can be reduced by queuing.

\section{Hardware Implementation}
\label{sec:hardware}

Figure~\ref{fig:architectures} shows the architecture
of EIE. 
A Central Control Unit (CCU) controls an array of PEs that
each computes one slice of the compressed network.
The CCU also receives non-zero input activations from a 
distributed {\em leading non-zero detection} network and
broadcasts these to the PEs.

Almost all computation in EIE is local to the PEs except for the collection of non-zero input activations that are broadcast to 
all PEs. However, the timing of the activation collection and
broadcast is non-critical as most PEs take many cycles to consume each input activation.

\textbf{Activation Queue and Load Balancing.} 
Non-zero elements of the input activation vector $a_j$
and their corresponding index $j$ are broadcast by the CCU to
an activation queue in each PE.
The broadcast is disabled if any PE has a full queue.
At any point in time each PE processes the
activation at the head of its queue.

The activation queue allows each PE to build up a backlog
of work to even out load imbalance that may
arise because the number of non zeros in a given column 
$j$ may vary from PE to PE.
In Section~\ref{sec:experimental} we measure the sensitivity of performance to the depth of the activation queue.

\textbf{Pointer Read Unit.}
The index $j$ of the entry at the head of the activation 
queue is used to look up the start and end pointers
$p_j$ and $p_{j+1}$ for the $v$ and $x$ arrays for
column $j$.
To allow both pointers to be read in one cycle using
single-ported SRAM arrays, we store pointers in 
two SRAM banks and use the LSB of the address to 
select between banks.
$p_j$ and $p_{j+1}$ will always be in different banks.
EIE pointers are 16-bits in length.

\textbf{Sparse Matrix Read Unit.} 
The sparse-matrix read unit uses
pointers $p_j$ and $p_{j+1}$ 
to read the non-zero 
elements (if any) of this PE's slice of column $I_j$
from the sparse-matrix SRAM.
Each entry in the SRAM is 8-bits in length and contains
one 4-bit element of $v$ and one 4-bit element of $x$.

For efficiency (see Section \ref{sec:experimental}) the PE's slice of encoded sparse matrix $I$ is stored in a 64-bit-wide SRAM.  
Thus eight entries are fetched on each SRAM read.
The high 13 bits of the current pointer $p$ selects
an SRAM row, and the low 3-bits select one of the eight
entries in that row.
A single $(v,x)$ entry is provided to the arithmetic unit each cycle.

\textbf{Arithmetic Unit.} 
The arithmetic unit receives a $(v,x)$ entry from the 
sparse matrix read unit and performs the multiply-accumulate 
operation $b_x = b_x + v \times a_j$.
Index $x$ is used to index an accumulator array (the destination activation registers) while $v$ is multiplied by the activation value at the head of the
activation queue.
Because $v$ is stored in 4-bit encoded form, it is
first expanded to a 16-bit fixed-point number via
a table look up.
A bypass path is provided to route the output of the adder
to its input if the same accumulator is selected on 
two adjacent cycles.

\textbf{Activation Read/Write.} 
The Activation Read/Write Unit contains two activation register files that accommodate the source and destination activation values respectively during a single round of FC layer computation. The source and destination register files exchange their role for next layer. Thus no additional data transfer is needed to support multi-layer feed-forward computation.

Each activation register file holds 64 16-bit activations.
This is sufficient to accommodate 4K activation vectors across 64 PEs.
Longer activation vectors can be accommodated with the 2KB activation SRAM. When the activation vector has a length greater than 4K, the M$\times$V will be completed in several batches, where each batch is of length 4K or less. All the local reduction is done in the register file. The SRAM is read only at the beginning and written at the end of the batch.

\textbf{Distributed Leading Non-Zero Detection.}
Input activations are hierarchically distributed to each PE. 
To take advantage of the input vector sparsity, we use leading non-zero detection logic to select the first non-zero result. 
Each group of 4 PEs does a local leading non-zero detection on their input activation. The result is sent to a Leading Non-zero Detection Node (LNZD Node) illustrated in Figure \ref{fig:architectures}.  Each LNZD node finds the next non-zero activation across its four children and sends this result up the quadtree. The quadtree is arranged so that wire lengths remain constant as we add PEs. At the root LNZD Node, the selected non-zero activation is broadcast back to all the PEs via a separate wire placed in an H-tree.

\textbf{Central Control Unit.}
The Central Control Unit (CCU) is the root LNZD Node. It communicates with the master, for example a CPU, and monitors the state of every PE by setting the control registers. There are two modes in the Central Unit: I/O and Computing. In the I/O mode, all of the PEs are idle while the activations and weights in every PE can be accessed by a DMA connected with the Central Unit. This is one time cost. In the Computing mode, the CCU repeatedly collects a non-zero value from the LNZD quadtree and broadcasts this value to all PEs. This process continues until the input length is exceeded. By setting the input length and starting address of pointer array, EIE is instructed to execute different layers.

\begin{figure}[t!]
\centering
	\includegraphics[width=0.27\textwidth]{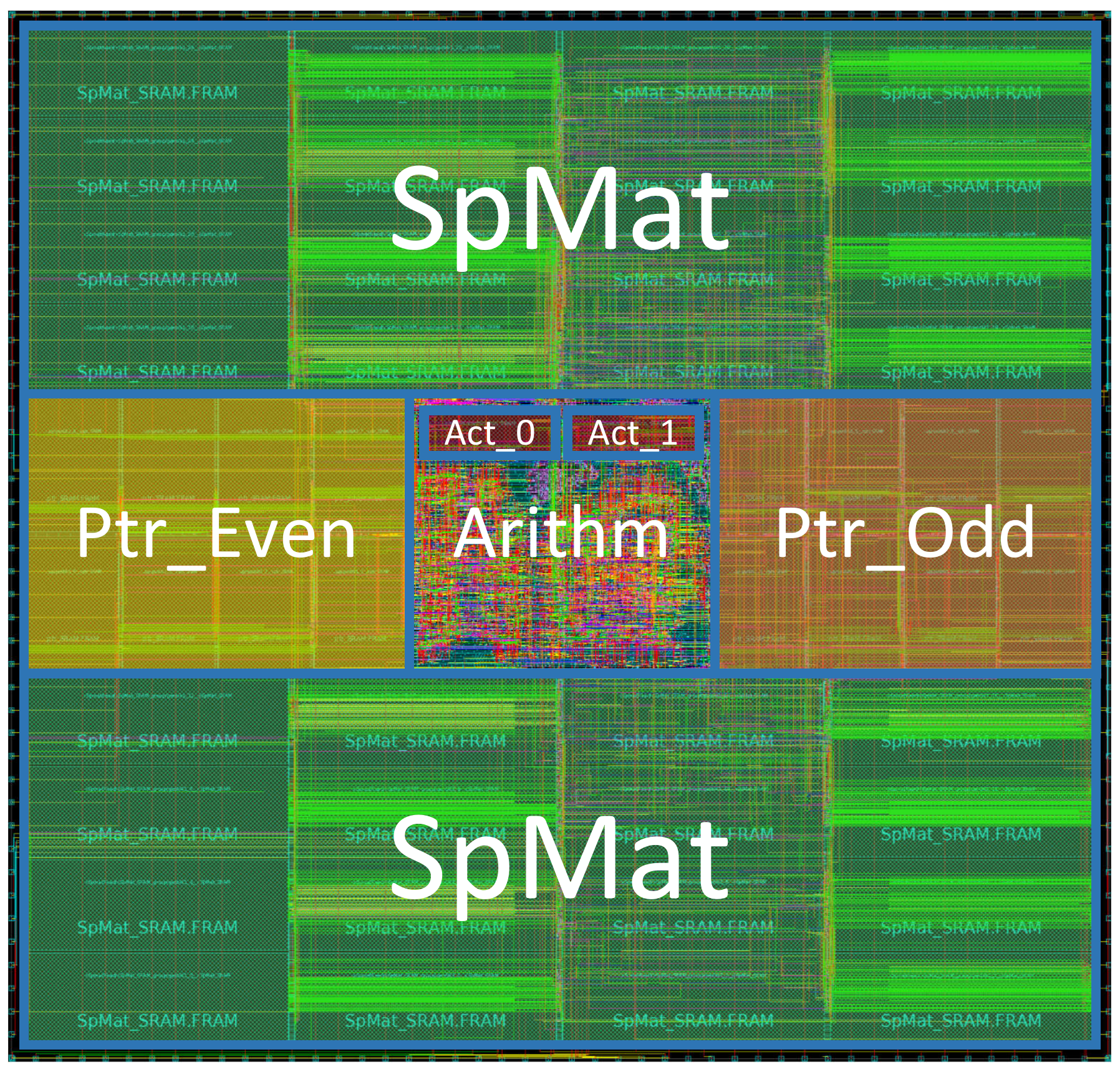}
    \vspace{-5pt}
	\caption{Layout of one PE in EIE under TSMC 45nm process.}
	\label{fig:layout}
\end{figure}

\begin{table}[t!]
	\footnotesize
	\centering	
        \caption{The implementation results of one PE in EIE and the breakdown by component type (line 3-7), by module (line 8-13). The critical path of EIE is 1.15 ns}
            \vspace{-10pt}
		\begin{tabular}{lrrrr}
		\hline
        \hline
		& \textbf{Power} & \multirow{2}{*}{\textbf{(\%)}} & \textbf{Area} & \multirow{2}{*}{\textbf{(\%)}} \\
		& \textbf{(mW)} &  &  \textbf{($\bm{\mu m^2}$)} & \\
        \hline
        Total & 9.157 &  & 638,024 &  \\		
        \hline
		memory & 5.416 & (59.15\%) & 594,786 & (93.22\%) \\
		clock network & 1.874 & (20.46\%) & 866 & (0.14\%) \\
		register & 1.026 & (11.20\%) & 9,465 & (1.48\%) \\
		combinational & 0.841 & (9.18\%) & 8,946 & (1.40\%) \\
		filler cell & & & 23,961 & (3.76\%) \\
		\hline
		Act\_queue & 0.112 & (1.23\%) & 758 & (0.12\%) \\
		PtrRead & 1.807 & (19.73\%) & 121,849 & (19.10\%) \\
		SpmatRead & 4.955 & (54.11\%) & 469,412 & (73.57\%) \\
		ArithmUnit & 1.162 & (12.68\%) & 3,110 & (0.49\%) \\
		ActRW & 1.122 & (12.25\%) & 18,934 & (2.97\%) \\
		filler cell &  &  & 23,961 & (3.76\%) \\		
		\hline	
        \hline
	\end{tabular}

    \vspace{-10pt}
	\label{table:breakdown}
\end{table}
\section{Evaluation Methodology}
\label{sec:evaluation}

\textbf{Simulator, RTL and Layout.} We implemented a custom cycle-accurate C++ simulator for the accelerator aimed to model the RTL behavior of synchronous circuits. Each hardware module is abstracted as an object that implements two abstract methods: propagate and update, corresponding to combination logic and the flip-flop in RTL. The simulator is used for design space exploration. It also serves as a checker for RTL verification. 

\begin{figure*}[!t]
  \vspace{-15pt}
\centering
	\includegraphics[width=0.95\textwidth]{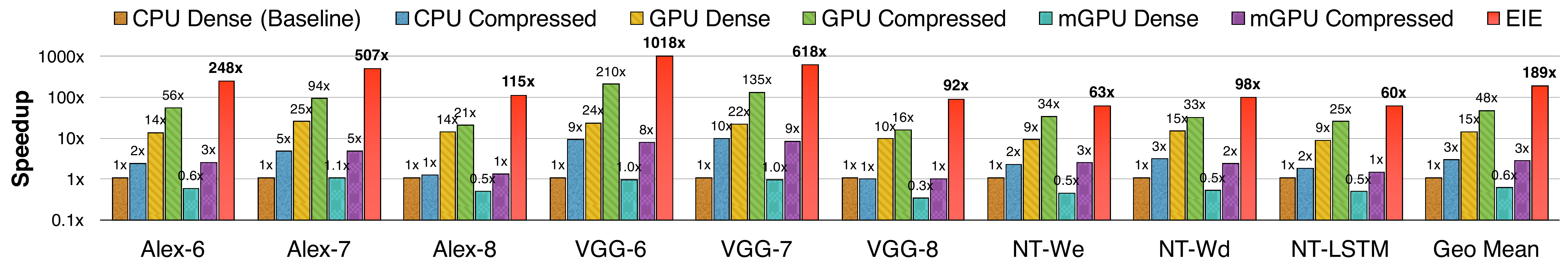}
    \vspace{-5pt}
	\caption{Speedups of GPU, mobile GPU and EIE compared with CPU running uncompressed DNN model. There is no batching in all cases. }
	\vspace{0pt}
	\label{fig:speedup}
\end{figure*}

\begin{figure*}[!t]
  \vspace{0pt}
\centering
	\includegraphics[width=0.96\textwidth]{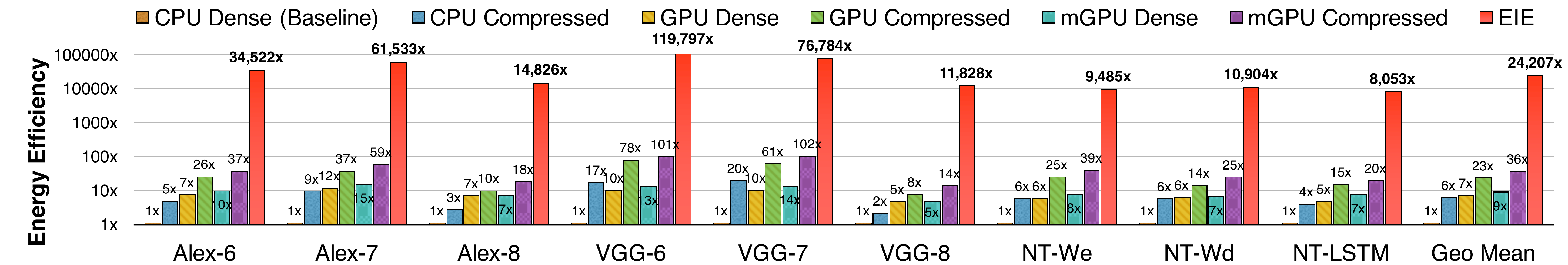}
    \vspace{-5pt}
	\caption{Energy efficiency of GPU, mobile GPU and EIE compared with CPU running uncompressed DNN model. There is no batching in all cases.}
    \vspace{-10pt}
	\label{fig:energy}
\end{figure*}

To measure the area, power and critical path delay, we implemented the RTL of EIE in Verilog. The RTL is verified against the cycle-accurate simulator. Then we synthesized EIE using the Synopsys Design Compiler~(DC) under the TSMC 45nm GP standard VT library with worst case PVT corner. We placed and routed the PE using the Synopsys IC compiler~(ICC). We used Cacti~\cite{cacti} to get SRAM area and energy numbers. We annotated the toggle rate from the RTL simulation to the gate-level netlist, which was dumped to switching activity interchange format~(SAIF), and estimated the power using Prime-Time PX.

\textbf{Comparison Baseline.} We compare EIE with three different off-the-shelf computing units: CPU, GPU and mobile GPU.

\textit{1) CPU.} We use Intel Core i-7 5930k CPU, a Haswell-E class processor, that has been used in NVIDIA Digits Deep Learning Dev Box as a CPU baseline.
To run the benchmark on CPU, we used MKL CBLAS GEMV to implement the original dense model and MKL SPBLAS CSRMV for the compressed sparse model.   
CPU socket and DRAM power are as reported by the \texttt{pcm-power} utility provided by Intel.

\textit{2) GPU.} We use NVIDIA GeForce GTX Titan X GPU, a state-of-the-art GPU for deep learning as our baseline using \texttt{nvidia-smi} utility to report the power.
To run the benchmark, we used cuBLAS GEMV to implement the original dense layer. For the compressed sparse layer, we stored the sparse matrix in in CSR format, and used cuSPARSE CSRMV kernel, which is optimized for sparse matrix-vector multiplication on GPUs.

\textit{3) Mobile GPU.} We use NVIDIA Tegra K1 that has 192 CUDA cores as our mobile GPU baseline. We used cuBLAS GEMV for the original dense model and cuSPARSE CSRMV for the compressed sparse model. Tegra K1 doesn't have software interface to report power consumption, so we measured the total power consumption with a power-meter, then assumed $15\%$ AC to DC conversion loss, $85\%$ regulator efficiency and $15\%$ power consumed by peripheral components~\cite{TK1,whitepaper} to report the AP+DRAM power for Tegra K1.

\begin{table}[h]
	\centering
	\footnotesize
	\caption{Benchmark from state-of-the-art DNN models}
	\vspace{-5pt}
	\begin{tabular}{p{1.1cm}|p{0.6cm}p{0.9cm}p{0.6cm}p{0.9cm}|l}
		\hline
		Layer & Size & Weight\% & Act\% & FLOP\% & Description  \\
		\hline
		\multirow{2}{*}{Alex-6} & 9216, & \multirow{2}{*}{9\%} & \multirow{2}{*}{35.1\%} & \multirow{2}{*}{3\%} & \multirow{3}{*}{Compressed} \\ 
		& 4096 & & & &\multirow{3}{*}{AlexNet~\cite{hinton12} for}  \\
		\cline{1-5}
		\multirow{2}{*}{Alex-7} & 4096, & \multirow{2}{*}{9\%} & \multirow{2}{*}{35.3\%} & \multirow{2}{*}{3\%} & \multirow{3}{*}{large scale image}  \\ 
		& 4096 & & & & \multirow{3}{*}{classification}\\
		\cline{1-5}
		\multirow{2}{*}{Alex-8} & 4096, & \multirow{2}{*}{25\%} & \multirow{2}{*}{37.5\%} & \multirow{2}{*}{10\%} & \\ 
		& 1000 & & & & \\
		\hline
		\multirow{2}{*}{VGG-6} & 25088, & \multirow{2}{*}{4\%} & \multirow{2}{*}{18.3\%} & \multirow{2}{*}{1\%} & \multirow{2}{*}{Compressed} \\ 
		& 4096 & & & & \multirow{2}{*}{VGG-16~\cite{simonyan2014very} for}  \\
		\cline{1-5}
		\multirow{2}{*}{VGG-7} & 4096, & \multirow{2}{*}{4\%} & \multirow{2}{*}{37.5\%} & \multirow{2}{*}{2\%} & \multirow{2}{*}{large scale image}  \\ 
		& 4096 & & & &\multirow{2}{*}{classification and}  \\
		\cline{1-5}
		\multirow{2}{*}{VGG-8} & 4096, & \multirow{2}{*}{23\%} & \multirow{2}{*}{41.1\%} & \multirow{2}{*}{9\%} & \multirow{2}{*}{object detection} \\ 
		& 1000 & & & & \\
		\hline
		\multirow{2}{*}{NT-We} & 4096, & \multirow{2}{*}{10\%} & \multirow{2}{*}{100\%} & \multirow{2}{*}{10\%} & Compressed \\ 
		& 600 & & & & NeuralTalk~\cite{neural_talk}  \\
		\cline{1-5}
		\multirow{2}{*}{NT-Wd} & 600, & \multirow{2}{*}{11\%} & \multirow{2}{*}{100\%} & \multirow{2}{*}{11\%} & with RNN and \\ 
		& 8791 & & & & LSTM for\\
		\cline{1-5}
		\multirow{2}{*}{NTLSTM} & 1201, & \multirow{2}{*}{10\%} &  \multirow{2}{*}{100\%} & \multirow{2}{*}{11\%} & automatic  \\ 
		& 2400 & & & & image captioning \\
		\hline
	\end{tabular}	
    \vspace{-10pt}
	\label{table2}
\end{table}

\begin{table*}[t!]
\vspace{-15pt}
	\centering
	\footnotesize
	\caption{Wall clock time comparison between CPU, GPU, mobile GPU and EIE. Unit: $\mu$s}
    \vspace{-5pt}
	\label{table:detailed_performance}
	\begin{tabular}{|l|l|l|l|l|l|l|l|l|l|l|l|}
		\hline
		\multirow{2}{*}{Platform} & Batch & Matrix & \multicolumn{3}{c|}{AlexNet} & \multicolumn{3}{c|}{VGG16} &  \multicolumn{3}{c|}{NT-} \\
		\cline{4-12}
		& Size & Type & FC6 & FC7 & FC8 & FC6 & FC7 & FC8 & We & Wd & LSTM \\
		\hline 	
		\multirow{2}{*}{CPU} & \multirow{2}{*}{1} & dense  & 7516.2 & 6187.1 & 1134.9 & 35022.8 & 5372.8 & 774.2 & 605.0 & 1361.4 & 470.5\\
		\cline{3-12}
		\multirow{2}{*}{(Core} & & sparse & 3066.5 & 1282.1 & 890.5 & 3774.3 & 545.1 & 777.3 & 261.2 & 437.4 & 260.0
		 \\
		\cline{2-12}
		\multirow{2}{*}{i7-5930k)}& \multirow{2}{*}{64} & dense & 318.4 & 188.9 & 45.8 & 1056.0 & 188.3 & 45.7 & 28.7 & 69.0 & 28.8 \\
		\cline{3-12}
		& & sparse & 1417.6 & 682.1 & 407.7 & 1780.3 & 274.9 & 363.1 & 117.7 & 176.4 & 107.4 \\
		\hline
		\multirow{3}{*}{GPU} & \multirow{2}{*}{1} & dense & 541.5 & 243.0 & 80.5 & 1467.8 & 243.0 & 80.5 & 65 & 90.1 & 51.9 \\
		\cline{3-12}
		\multirow{3}{*}{(Titan X)} & & sparse & 134.8 & 65.8 & 54.6 & 167.0 & 39.8 & 48.0 & 17.7 & 41.1 & 18.5 \\
		\cline{2-12}
		& \multirow{2}{*}{64} & dense   & 19.8 & 8.9 & 5.9 & 53.6 & 8.9 & 5.9 & 3.2 & 2.3 & 2.5 \\
		\cline{3-12}
		& & sparse & 94.6 & 51.5 & 23.2 & 121.5 & 24.4 & 22.0 & 10.9 & 11.0 & 9.0 \\
		\hline		
		\multirow{3}{*}{mGPU} & \multirow{2}{*}{1} & dense  & 12437.2 & 5765.0 & 2252.1 & 35427.0  & 5544.3  & 2243.1 & 1316 & 2565.5 & 956.9 \\
		\cline{3-12}
		\multirow{3}{*}{(Tegra K1)} & & sparse & 2879.3 & 1256.5 & 837.0 & 4377.2  & 626.3 & 745.1 & 240.6 & 570.6 & 315 \\
		\cline{2-12}
		& \multirow{2}{*}{64} & dense  & 1663.6 & 2056.8 & 298.0 & 2001.4 & 2050.7 & 483.9 & 87.8 & 956.3 & 95.2 \\		
		\cline{3-12}
		& & sparse & 4003.9 & 1372.8 & 576.7 & 8024.8 & 660.2 & 544.1 & 236.3 & 187.7 & 186.5 \\ 
		\hline
		\multirow{2}{*}{\textbf{EIE}} & \multicolumn{2}{c|}{\textbf{Theoretical Time}} & \textbf{28.1} & \textbf{11.7} & \textbf{8.9} & \textbf{28.1} & \textbf{7.9} & \textbf{7.3} & \textbf{5.2} & \textbf{13.0} & \textbf{6.5} \\
		\cline{2-12}
		& \multicolumn{2}{c|}{\textbf{Actual Time}} & \textbf{30.3} & \textbf{12.2} & \textbf{9.9} & \textbf{34.4} & \textbf{8.7} & \textbf{8.4} & \textbf{8.0} & \textbf{13.9} & \textbf{7.5} \\  
		\hline
	\end{tabular}
\end{table*}

\begin{figure*}[t!]
\centering
	\includegraphics[width=0.9\textwidth]{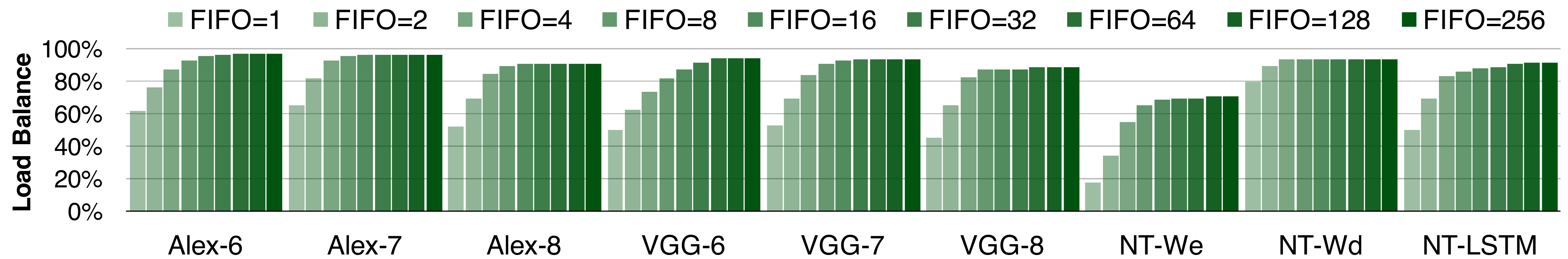}
    \vspace{-5pt}
	\caption{Load efficiency improves as FIFO size increases. When FIFO deepth$>$8, the marginal gain quickly diminishes. So we choose FIFO depth$=$8.}
        \vspace{-15pt}
	\label{fig:FIFO}
\end{figure*}
\textbf{Benchmarks.} We compare the performance on two sets of models: uncompressed DNN model and the compressed DNN model. The uncompressed DNN model is obtained from Caffe model zoo~\cite{caffe} and NeuralTalk model zoo~\cite{neural_talk}; The compressed DNN model is produced as described in~\cite{han2015deep,song_pruning}. The benchmark networks have 9 layers in total obtained from AlexNet, VGGNet, and NeuralTalk. We use the Image-Net dataset~\cite{imagenet} and the Caffe~\cite{caffe} deep learning framework as golden model to verify the correctness of the hardware design.

\section{Experimental Results}
\label{sec:experimental}
Figure~\ref{fig:layout} shows the layout (after place-and-route) of an EIE processing element. The power/area breakdown is shown in Table~\ref{table:breakdown}. 
We brought the critical path delay down to 1.15ns by introducing 4 pipeline stages to update one activation: codebook lookup and address accumulation (in parallel), output activation read and input activation multiply (in parallel), shift and add, and output activation write.
Activation read and write access a local register and activation bypassing is employed to avoid a pipeline hazard.  Using 64 PEs running at 800MHz yields a performance of 102 GOP/s. Considering $10\times$ weight sparsity and $3\times$ activation sparsity, this requires a dense DNN accelerator 3TOP/s to have equivalent application throughput.

The total SRAM capacity (Spmat+Ptr+Act) of each EIE PE is 162KB. The activation SRAM is 2KB storing activations. The Spmat SRAM is 128KB storing the compressed weights and indices.  Each weight is 4bits, each index is 4bits. Weights and indices are grouped to 8bits and addressed together. The Spmat access width is optimized at 64bits. The Ptr SRAM is 32KB storing the pointers in the CSC format. In the steady state, both Spmat SRAM and Ptr SRAM are accessed every $64/8=8$ cycles.
The area and power is dominated by SRAM, the ratio is 93\% and 59\% respectively. Each PE is $0.638mm^2$ consuming $9.157mW$.  Each group of 4 PEs needs a LNZD unit for nonzero detection. A total of 21 LNZD units are needed for 64 PEs ($16+4+1=21$).  Synthesized result shows that one LNZD unit takes only $0.023mW$ and an area of $189um^2$, less than 0.3$\%$ of a PE.

\subsection{Performance}

We compare EIE against CPU, desktop GPU and the mobile GPU on 9 benchmarks selected from AlexNet, VGG-16 and Neural Talk. The overall results are shown in Figure~\ref{fig:speedup}. 
There are 7 columns for each benchmark, comparing the computation time of EIE on compressed network over CPU~/~GPU~/~TK1 on uncompressed~/~compressed network. Time is normalized to CPU.
EIE significantly outperforms the general purpose hardware and is, on average, 189$\times$, 13$\times$, 307$\times$ faster than CPU, GPU and mobile GPU respectively.

EIE's theoretical computation time is calculated by dividing workload GOPs by peak throughput. The actual computation time is around 10\% more than the theoretical computation time due to load imbalance. In Fig.~\ref{fig:speedup}, the comparison with CPU~/~GPU~/~TK1 is reported using actual computation time. The wall clock time of CPU~/~GPU~/~TK1/~EIE for all benchmarks are shown in Table~\ref{table:detailed_performance}. 

\begin{figure*}[!t]
 \vspace{-15pt}
\begin{center}
\begin{tabular}{@{} c @{} c}
\includegraphics[width=.45\textwidth]{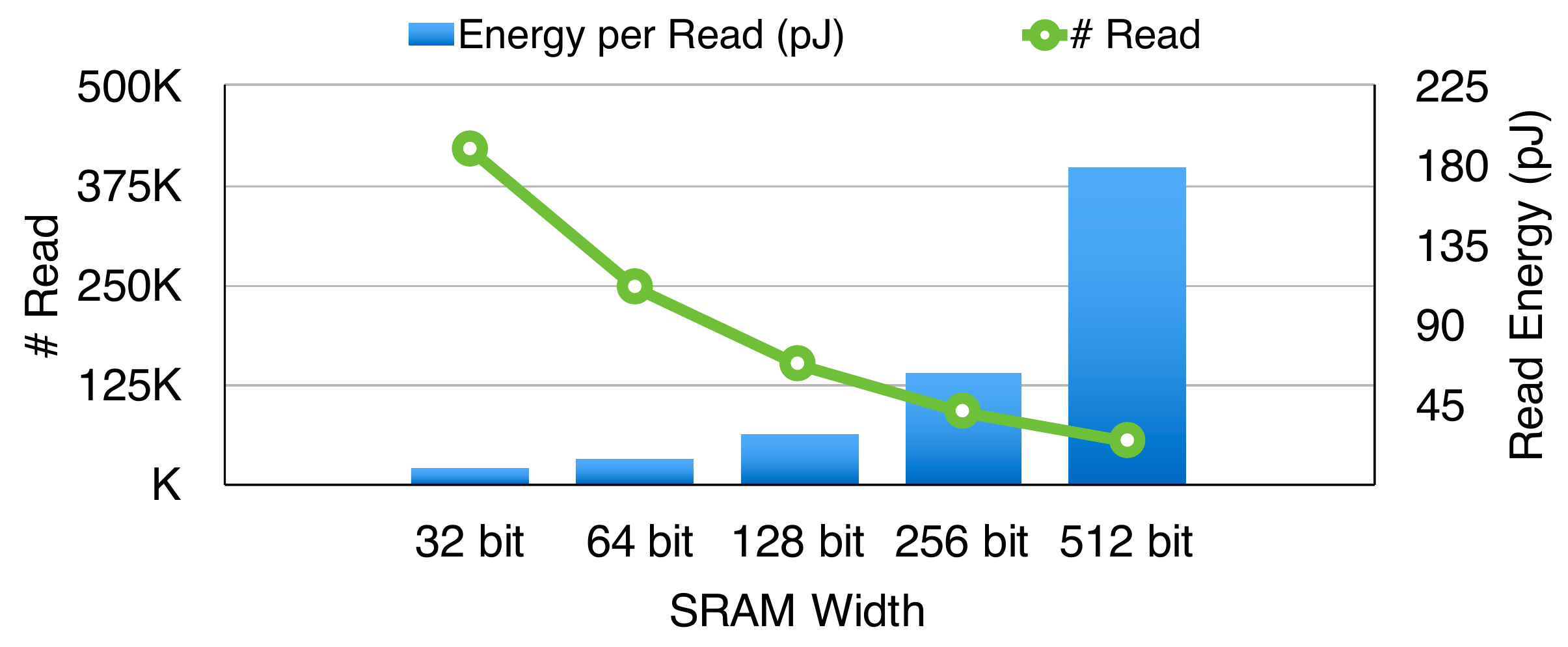}
& \hspace{10pt}
\includegraphics[width= .41\textwidth]{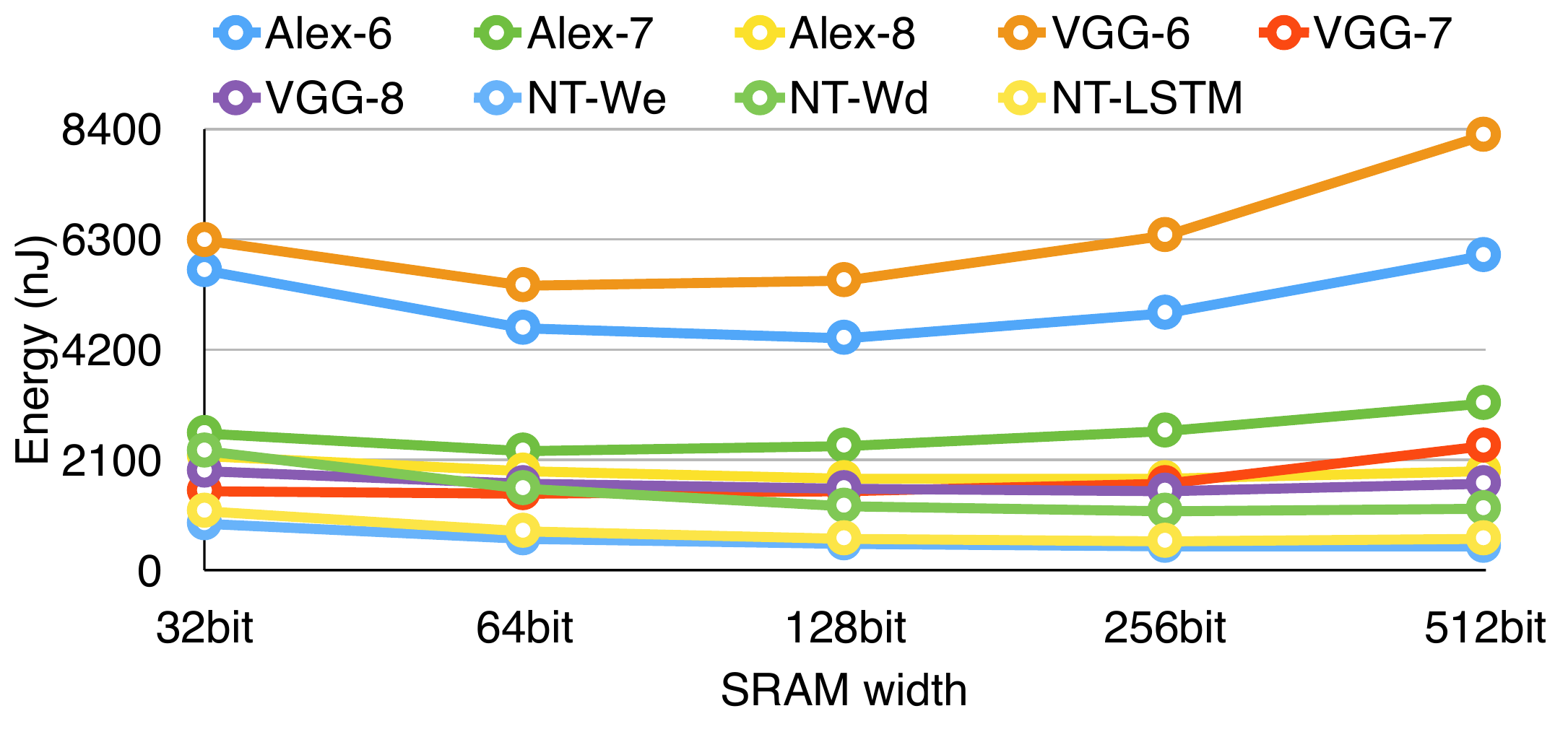}
\end{tabular}
\end{center}
 \vspace{-15pt}
\caption{Left: SRAM read energy and number of reads benchmarked on AlexNet. 
Right: Multiplying the two curves in the left gives the total energy consumed by SRAM read.}
\label{fig:SRAM_tradeoff}
 \vspace{-10pt}
\end{figure*}

EIE is targeting extremely latency-focused applications, which require real-time inference. Since assembling a batch adds significant amounts of latency, we consider the case when batch size = 1 when benchmarking the performance and energy efficiency with CPU and GPU as shown in Figure~\ref{fig:speedup}. As a comparison, we also provided the result for batch size = 64 in Table~\ref{table:detailed_performance}. EIE outperforms most of the platforms and is comparable to desktop GPU in the batching case. 

The GOP/s required for EIE to achieve the same application throughput (Frames/s) is much lower than competing approaches because EIE exploits sparsity to eliminate 97\% of the GOP/s performed by dense approaches. 3 TOP/s on an uncompressed network requires only 100 GOP/s on a compressed network. EIE's throughput is scalable to over 256 PEs. Without EIE's dedicated logic, however, model compression by itself applied on a CPU/GPU yields only $3\times$ speedup.

\subsection{Energy}

In Figure~\ref{fig:energy}, we report the energy efficiency comparisons of M$\times$V on different benchmarks. There are 7 columns for each benchmark, comparing the energy efficiency of EIE on compressed network over CPU~/~GPU~/~TK1 on uncompressed~/~compressed network. Energy is obtained by multiplying computation time and total measured power as described in section~\ref{sec:evaluation}.

EIE consumes on average,  $24,000\times$, $3,400\times$, and $2,700\times$ less energy compared to CPU, GPU and the mobile GPU respectively. This is a 3-order of magnitude energy saving from three places: first, the required energy per memory read is saved (SRAM over DRAM): using a compressed network model enables state-of-the-art neural networks to fit in on-chip SRAM, reducing energy consumption by 120$\times$ compared to fetching a dense uncompressed model from DRAM (Figure~\ref{fig:energy}). Second, the number of required memory reads is reduced. The compressed DNN model has 10\% of the weights where each weight is quantized by only 4 bits. Lastly, taking advantage of vector sparsity saved 65.14\% redundant computation cycles. Multiplying those factors $120\times10\times8\times3$ gives $28,800\times$ theoretical energy saving. Our actual savings are about $10\times$ less than this number because of index overhead and because EIE is implemented in 45nm technology compared to the 28nm technology used by the Titan-X GPU and the Tegra K1 mobile GPU.

\subsection{Design Space Exploration}

\textbf{Queue Depth.}
The activation FIFO queue deals with load imbalance between the PEs. A deeper FIFO queue can better decouple producer and consumer, but with diminishing returns, as shown in our experiment in Figure~\ref{fig:FIFO}. We varied the FIFO queue depth from from 1 to 256 in powers of 2 across 9 benchmarks using 64 PEs, and measured the load balance efficiency. This efficiency is defined as: 1 $-$ bubble cycles~(due to starvation) divided by total computation cycles. At FIFO size $=$ 1, around half of the total cycles are idle and the accelerator suffers from severe load imbalance. Load imbalance is reduced as FIFO depth is increased but with diminishing returns beyond a depth of 8. Thus, we choose 8 as the optimal queue depth. 

\begin{figure}[t]
\centering
	\includegraphics[width=0.45\textwidth]{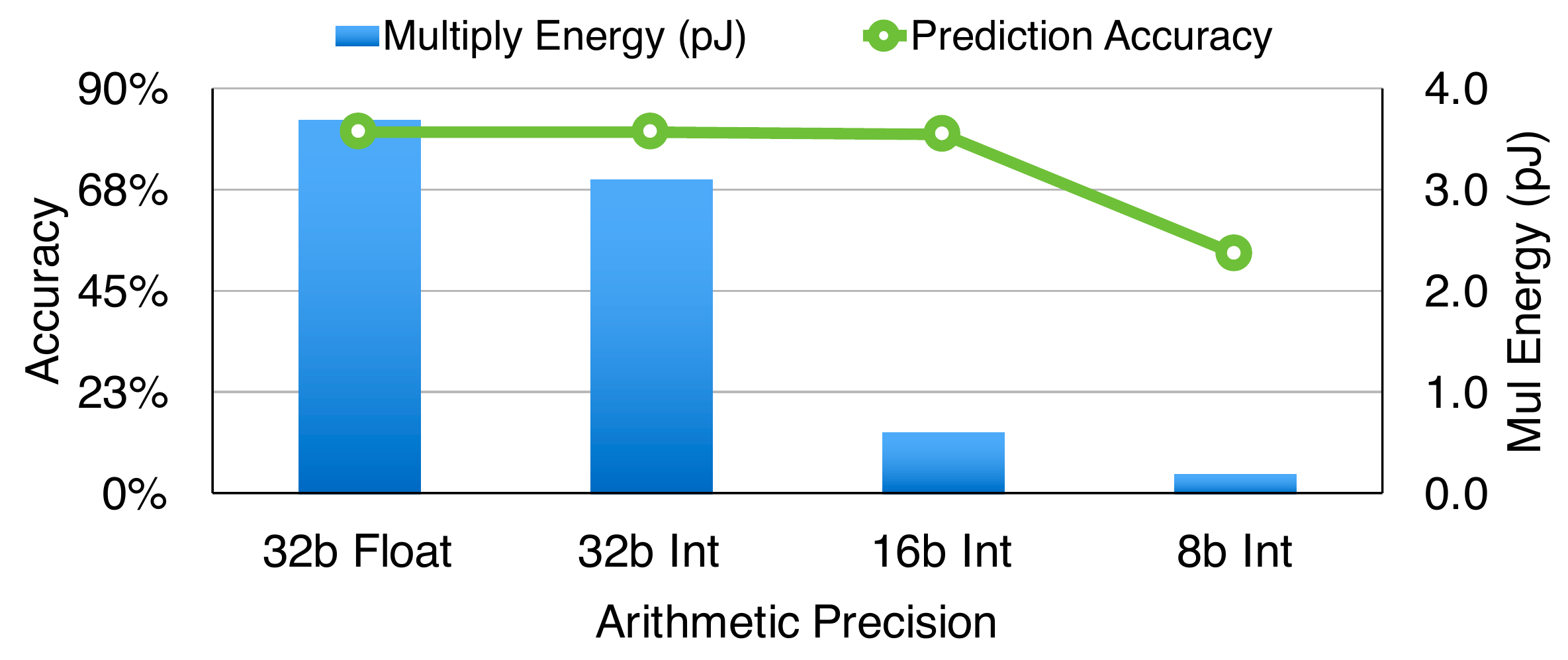}
        \vspace{-5pt}
	\caption{Prediction accuracy and multiplier energy with different arithmetic precision. }
	\label{fig:ALU}
    \vspace{-10pt}
\end{figure}
 
Notice the NT-We benchmark has poorer load balance efficiency compared with others. This is because that it has only 600 rows. Divided by 64 PEs and considering the 11\% sparsity, each PE on average gets a single entry, which is highly susceptible to variation among PEs, leading to load imbalance. Such small matrices are more efficiently executed on 32 or fewer PEs.
\begin{figure*}[!t]
\vspace{-15pt}
\centering
	\includegraphics[width=0.99\textwidth]{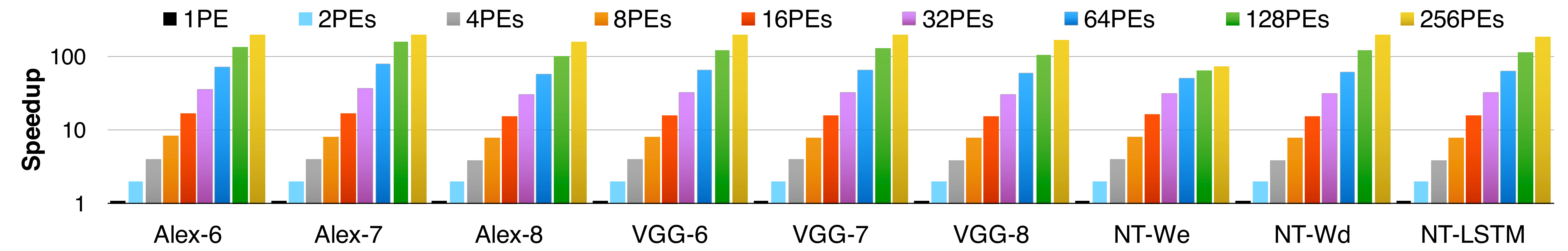}
\vspace{-5pt}
	\caption{System scalability. It measures the speedups with different numbers of PEs. The speedup is near-linear.}
	\label{fig:scalability}
    \vspace{-10pt}
\end{figure*}


\textbf{SRAM Width.} We choose an SRAM with a 64-bit interface to store the sparse matrix~(Spmat) since it minimized the total energy. Wider SRAM interfaces reduce the number of total SRAM accesses, but increase the energy cost per SRAM read. The experimental trade-off is shown in Figure~\ref{fig:SRAM_tradeoff}. SRAM energy is modeled using Cacti~\cite{cacti} under 45nm process. SRAM access times are measured by the cycle-accurate simulator on AlexNet benchmark. As the total energy is shown on the right, the minimum total access energy is achieved when SRAM width is 64 bits. 
For larger SRAM widths, read data is wasted: the typical number of activation elements of FC layer is 4K\cite{hinton12,simonyan2014very} so assuming 64 PEs and 10\% density~\cite{song_pruning}, each column in a PE will have 6.4 elements on average.  This matches a 64-bit SRAM interface that provides 8 elements. If more elements are fetched and the next column corresponds to a zero activation, those elements are wasted.

\textbf{Arithmetic Precision.} We use 16-bit fixed-point arithmetic. As shown in Figure~\ref{fig:ALU}, 16-bit fixed-point multiplication consumes $5\times$ less energy than 32-bit fixed-point and $6.2\times$ less energy than 32-bit floating-point. At the same time, using 16-bit fixed-point arithmetic results in less than 0.5\% loss of prediction accuracy: 79.8\% compared with 80.3\% using 32-bit floating point arithmetic. With 8-bits fixed-point, however, the accuracy dropped to only 53\%, which becomes intolerable. The accuracy is measured on ImageNet dataset~\cite{imagenet} with AlexNet~\cite{hinton12}, and the energy is obtained from synthesized RTL under 45nm process.

`Deep Compression' does not affect accuracy\cite{song_pruning}\cite{han2015deep}\cite{SqueezeNet}, but using 16-bit arithmetic degrades accuracy by only 0.5\%. In order to have absolute no loss of accuracy, switching to 32 bit arithmetic would not substantially affect the power or area of EIE. Only the 16-entry codebook, the arithmetic units, and the activation register files would change in width. The index stored in SRAM would remain 4-bits. The majority of power/area is taken by SRAM, not arithmetic.  The area used by filler cells (used to fill blank area) is sufficient to double the area of the arithmetic units and activation registers (Table~\ref{table:breakdown}).

\section{Discussion}
\label{sec:discussion}

Many engines have been proposed for Sparse Matrix-Vector multiplication~(SPMV) and the existing trade-offs on the targeted platforms are studied~\cite{SpMXV:FPGA:2005,SpMxV:FPGA:2014}.
There are typically three approaches to partition the workload for matrix-vector multiplication. The combination of these methods with storage format of the Matrix creates a design space trade-off. 

\subsection{Workload Partitioning}
The first approach is to distribute matrix columns to PEs. Each PE handles the multiplication between its columns of $W$ and corresponding element of $a$ to get a partial sum of the output vector $b$. 
The benefit of this solutions is that each element of $a$ is only associated with one PE --- giving full locality for vector $a$. The drawback is that a reduction operation between PEs is required to obtain the final result. 

A second approach (ours) is to distribute matrix rows to PEs. A central unit broadcasts one vector element $a_j$ to all PEs. Each PE computes a number of output activations $b_i$ by performing inner products of the corresponding row of $W$, $W_j$ that is stored in the PE  with vector $a$. 
The benefit of this solutions is that each element of $b$ is only associated with one PE --- giving full locality for vector $b$. The drawback is that vector $a$ needs to be broadcast to all PEs.

A third approach combines the previous two approaches by distributing blocks of $W$ to the PEs in 2D fashion. This solution is more scalable for distributed systems where communication latency cost is significant~\cite{eijkhout1992lapack}. This way both of the collective communication operations "Broadcast" and "Reduction" are exploited but in a smaller scale and hence this solution is more scalable.

The nature of our target class of application and its sparsity pattern affects the constraints and therefore our choice of partitioning and storage. The density of $W$ is  $\approx 10\%$, and the density of $a$ is $\approx 30\%$, both with random distribution. Vector $a$ is stored in normal dense format and contains $70\%$ the zeros in the memory, because for different input, $a_j$'s sparsity pattern differs. We want to utilize the sparsity of both  $W$ and $a$.

The first solution suffers from load imbalance given that vector $a$ is also sparse. Each PE is responsible for a column. PE$_j$ will be completely idle if their corresponding element $a_j$ is zero. On top of the Idle PEs, this solution requires across-PE reduction and extra level of synchronization. 

Since the SPMV engine, has a limited number of PEs, there won't be a scalability issue to worry about. However, the hybrid solution will suffer from inherent complexity and still possible load imbalance since multiple PEs sharing the same column might remain idle. 

We build our solution based on the second distribution scheme taking the 30\% density of vector $a$ into account. Our solution aims to perform computations by in-order look-up of nonzeros in $a$. Each PE gets all the non-zero elements of $a$ in order and performs the inner products by looking-up the matching element that needs to be multiplied by $a_j$, $W_j$. This requires the matrix $W$ being stored in CSC format so the PE can multiply all the elements in the $j$-th column of $W$ by $a_j$.

\subsection{Scalability}

As the matrix gets larger, the system can be scaled up by adding more PEs. Each PE has local SRAM storing distinct rows of the matrix without duplication, so the SRAM is efficiently utilized. 

Wire delay increases with the square root of the number of PEs, however, this is not a problem in our architecture. Since EIE only requires one broadcast over the computation of the entire column, which takes many cycles.
Consequently, the broadcast is not on the critical path and can be pipelined because FIFOs decouple producer and consumer.

Figure~\ref{fig:scalability} shows EIE achieves good scalability on all benchmarks except NT-We. NT-We is very small ($4096\times600$). Dividing the columns of size 600 and sparsity 10\% to 64 or more PEs causes serious load imbalance.

\begin{figure*}[!t]
\vspace{-15pt}
\centering
	\includegraphics[width=0.9\textwidth]{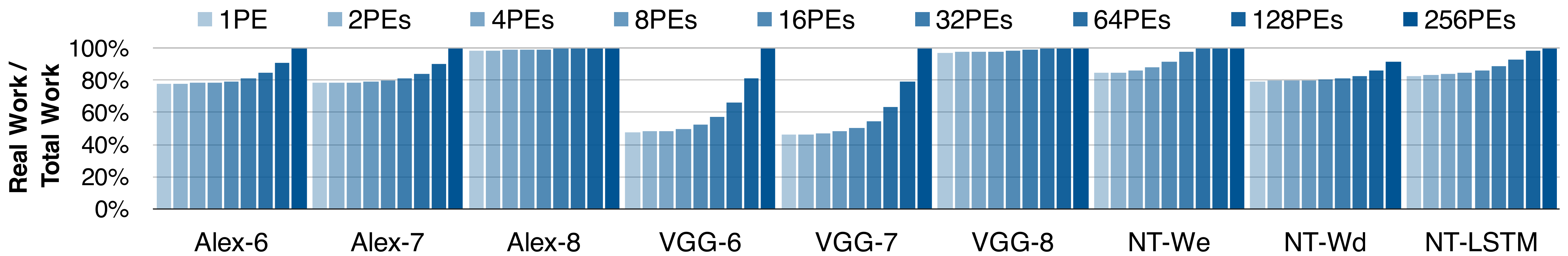}
    \vspace{-5pt}
	\caption{As the number of PEs goes up, the number of padding zeros decreases, leading to less padding zeros and less redundant work, thus better compute efficiency. }
	\label{fig:padding}
\vspace{-5pt}
\end{figure*}

\begin{figure*}[!t]
\centering
	\includegraphics[width=0.9\textwidth]{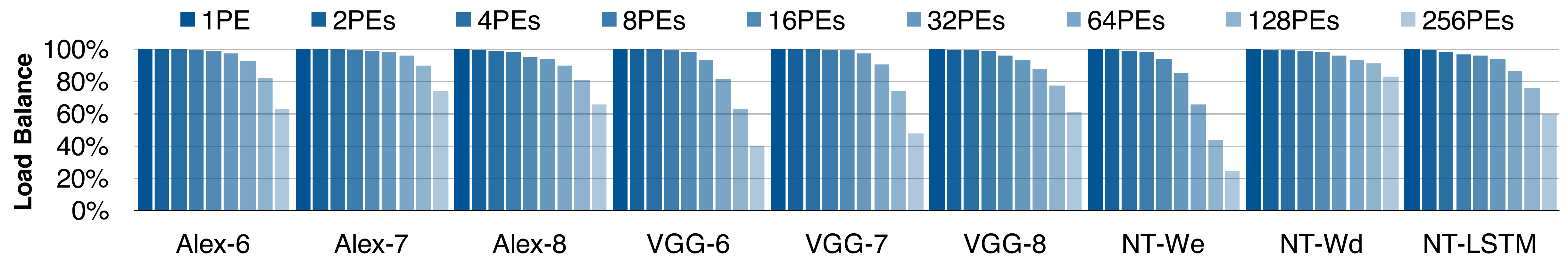}
    \vspace{-5pt}
	\caption{Load efficiency is measured by the ratio of stalled cycles over total cycles in ALU.  More PEs lead to worse load balance, but less padding zeros and more useful computation.}
\label{fig:load_balance}
\vspace{-10pt}
\end{figure*}

Figure~\ref{fig:padding} shows the number of padding zeros with different number PEs. Padding zero occur when the jump between two consecutive non-zero element in the sparse matrix is larger than 16, the largest number that 4 bits can encode. Padding zeros are considered non-zero and lead to wasted computation. Using more PEs reduces padding zeros, because  the  distance between non-zero elements get smaller due to matrix partitioning, and 4-bits encoding a max distance of 16 will more likely be enough.

Figure~\ref{fig:load_balance} shows the load balance with different number of PEs, measured with FIFO depth equal to 8. With more PEs, load balance becomes worse, but padding zero overhead decreases, which yields efficiency for most benchmarks remain constant. The scalability result is plotted in figure \ref{fig:scalability}.

\subsection{Flexibility}

EIE is designed for large neural networks. The weights and input/ouput of most layers can be easily fit into EIE's storage. For those with extremely large input/output sizes (for example, FC6 layer of VGG-16 has an input size of 25088), EIE is still able to execute them with 64PEs.

EIE can assist general-purpose processors for sparse neural network acceleration or other tasks related to SPMV. One type of neural network structure can be decomposed into certain control sequence so that by writing control sequence to the registers of EIE, a network could be executed.

EIE has the potential to support 1x1 convolution and 3x3 Winograd convolution by turning the channel-wise reduction into an $M\times V$. Winograd convolution saves $2.25\times$ multiplications than naive convolution\cite{lavin2015fast}, and for each Winograd patch the 16 $M\times V$ can be scheduled on an EIE.


\section{Comparison with Related Work}
\label{sec:related}

We compare the results of performance, power and area on $M\times V$ in Table~\ref{tab:comparsion_platform}. The performance is evaluated on the FC7 layer of AlexNet\footnote{Except for TrueNorth where FC7 result is not provided, using TIMIT LSTM result for comparison instead~(different benchmarks differ $<2\times$).}. We compared six platforms for neural networks: Core-i7~(CPU), Titan X~(GPU), Tegra K1~(mobile GPU), A-Eye~(FPGA), DaDianNao~(ASIC), TrueNorth~(ASIC). All other four platforms suffer from low-efficiency during matrix-vector multiplication. A-Eye is optimized for CONV layers and all of the parameters are fetched from the external DDR3 memory, making it extremely sensitive to bandwidth problem. DaDianNao distributes weights on 16 tiles, each tile with 4 eDRAM banks, thus has a peak memory bandwidth of $16\times 4\times (1024bit/8)\times606MHz=4964 GB/s$. Its performance on $M\times V$ is estimated based on the peak memory bandwidth because $M\times V$ is completely memory bound. In contrast, EIE maintains a high throughput for $M\times V$ because after compression, all weights fit in on-chip SRAM, even for very large-scale networks. With 256PEs, EIE has $3.25\times$ more throughput than 64PEs and can hold 336 million parameters, even larger than VGGnet. 
The right column projected EIE to the same technology (28nm) as the other platforms, with 256PEs, EIE has $2.9\times$ throughput, $3\times$ area efficiency and $19\times$ power efficiency than DaDianNao. 

\textbf{Model Compression.}
DNN model compression is widely used to reduce the storage required by DNN models. In early work, network pruning proved to be a promising approach to reducing the network complexity and over-fitting \cite{hanson1989comparing, lecun1989optimal, hassibi1993second}. Recently Han et al.~\cite{song_pruning} pruned connections in large scale neural networks and achieved $9\times$ and $13
\times$ pruning rate for AlexNet and VGG-16 with no loss of accuracy on ImageNet. The followup work `Deep Compression'~\cite{han2015deep} compressed DNN by pruning, weight sharing and Huffman coding, pushing the compression ratio to 35-49$\times$. SqueezeNet~\cite{SqueezeNet} further pushes this ratio to 510$\times$: it combines new ConvNet architecture and Deep Compression, its model size is only 470KB-660KB and has the same prediction accuracy as AlexNet. 
SqueezeNet makes it easy to fit DNN model fully in SRAM and is available online\footnote{The 660KB version of SqueezeNet model trained on Imagenet is available at: http://songhan.github.io/SqueezeNet-Deep-Compression}. SVD is frequently used to reduce model size~\cite{denton2014LinearStructure}\cite{zhang2014efficient}. Minerva~\cite{minerva} also uses data quantization to save memory energy. 

Model compression is crucial for reducing memory energy. However, model compression by itself applied on a GPU yields only $3\times$ energy savings\cite{han2015deep}, while EIE increases this to $3000\times$ by tailoring an architecture to exploit the irregularity and decoding created by model compression.

\textbf{DNN accelerator.}
Many custom accelerators have been proposed for DNNs. 
DianNao~\cite{diannao} implements an array of multiply-add units to map large DNN onto its core architecture. Due to limited SRAM resource, the off-chip DRAM traffic dominates the energy consumption.
DaDianNao~\cite{dadiannao} and ShiDianNao~\cite{shidiannao} eliminate the DRAM access by having all weights on-chip (eDRAM or SRAM).

\begin{savenotes}
\begin{table*}[t!]
	\centering	
        \vspace{-15pt}
    \caption{Comparison with existing hardware platforms for DNNs.}
    \label{tab:comparsion_platform}
    \vspace{-5pt}
	\begin{tabularx}{\textwidth}{|l|X|X|X|X|X|X|X|X|}
		\hline
		Platform & Core-i7 5930K & GeForce Titan X & Tegra K1 & A-Eye \cite{fpga2016cnn} & Da-DianNao\cite{dadiannao} & True- North\cite{esser2016convolutional} & \bf{EIE (ours, 64PE)} & \bf{EIE (28nm, 256PE)}  \\ 
		\hline
		Year & 2014 & 2015 & 2014 & 2015 & 2014 & 2014 & 2016 & 2016 \\											\hline
		Platform Type & CPU & GPU & mGPU & FPGA & ASIC & ASIC & ASIC & ASIC \\
		\hline
		Technology & 22nm & 28nm & 28nm & 28nm & 28nm & 28nm & 45nm & 28nm \\
		\hline
        Clock ({\em MHz}) & 3500 &  1075 & 852 & 150 & 606 & Async & 800 & 1200 \\
        \hline
		Memory type & DRAM & DRAM & DRAM & DRAM & eDRAM & SRAM & SRAM & SRAM \\
		\hline
		Max DNN model size ({\it \#Params}) & $<$16G & $<$3G & $<$500M & $<$500M & 18M & 256M & 84M & 336M \\
		\hline
		Quantization Stategy & 32-bit float & 32-bit float & 32-bit float & 16-bit fixed & 16-bit fixed & 1-bit fixed & 4-bit fixed & 4-bit fixed \\																							

		\hline
        Area (\em{mm\textsuperscript{2}}) & 356 & 601 & - & - & 67.7 & 430 & 40.8 & 63.8 \\       
		\hline
		Power~(\em{W}) & 73 & 159 & 5.1 & 9.63 & 15.97 & 0.18 & 0.59 & 2.36 \\
        \hline		
		M$\times$V Throughput~({\em Frames/s}) & 162 & 4,115 & 173 & 33 & 147,938 & 1,989 & 81,967 & 426,230 \\
		\hline
	    Area Efficiency~( \em{Frames/s/mm\textsuperscript{2}}) & 0.46 & 6.85 & - & - & 2,185 & 4.63 & 2,009 & 6,681 \\
        \hline
		Energy Efficiency~({\em Frames/J}) & 2.22 & 25.9 & 33.9 & 3.43 & 9,263 & 10,839 & 138,927 & 180,606\\		
		\hline
	\end{tabularx}
    \vspace{-10pt}
\end{table*}
\end{savenotes}

In both architectures, the weights are uncompressed and stored in the dense format. As a result, ShiDianNao can only handle very small DNN models up to 64K parameters, which is 3 orders of magnitude smaller than the 60 Million parameter AlexNet by only containing 128KB on-chip RAM. Such large networks are impossible to fit on chip on ShiDianNao without compression.

DaDianNao stores the uncompressed model in eDRAM taking 6.12W memory power and 15.97W total power. EIE stores compressed model in SRAM taking only 0.35W memory power and only 0.59W total power; DaDianNao cannot exploit the sparsity from weights and activations and they must expand the network to dense form before operation. It can not exploit weight sharing either. Using just the compression (and decompressing the data before computation) would reduce DaDianNao total power to around 10W in 28nm, compared to EIE’s power of 0.58W in 45nm.

Previous DNN accelerators targeting ASIC and FPGA platforms~\cite{diannao}\cite{zhang2015optimizing} used mostly CONV layer as benchmarks, but have few dedicated experiments on FC layers, which has significant bandwidth bottlenecks, and is widely used in RNN and LSTMs. Loading weights from the external memory for the FC layer may significantly degrade the overall performance of the network\cite{fpga2016cnn}.

\textbf{Sparse Matrix-Vector Multiplication Accelerator.}
There is research effort on the implementation of sparse matrix-vector multiplication (SPMV) on general-purpose processors.  Monakov et al.~\cite{Automatically:GPU:Tuning:SMVM} proposed a matrix storage format that improves locality, which has low memory footprint and enables automatic parameter tuning on GPU. Bell et al.~\cite{2008:NVIDIA:tech:bell} implemented data structures and algorithms for SPMV on GeForce GTX 280 GPU and achieved performance of 36 GFLOP/s in single precision and 16 GFLOP/s in double precision. \cite{Throughput:oriented:Processors:SC:09} developed SPMV techniques that utilizes large percentages of peak bandwidth for throughput-oriented architectures like the GPU. They achieved over an order of magnitude performance improvement over a quad-core Intel Clovertown system.

To pursue a better computational efficiency, several recent works focus on using FPGA as an accelerator for SPMV.
Zhuo et al.~\cite{SpMXV:FPGA:2005} proposed an FPGA-based design on Virtex-II Pro for SPMV. Their design outperforms general-purpose processors, but the performance is limited by memory bandwidth.
Fowers et al.~\cite{SMVM:FCCM:2014} proposed a novel sparse matrix encoding and an FPGA-optimized architecture for SPMV. With lower bandwidth, it achieves $2.6\times$ and $2.3\times$ higher power efficiency over CPU and GPU respectively while having lower performance due to lower memory bandwidth.
Dorrance et al.~\cite{SpMxV:FPGA:2014} proposed a scalable SMVM kernel on Virtex-5 FPGA. It outperforms CPU and GPU counterparts with$>$$300\times$ computational efficiency and has 38-50$\times$ improvement in energy efficiency.

For compressed deep networks, previously proposed SPMV accelerators can only exploit the static weight sparsity. They are unable to exploit dynamic activation sparsity~($3\times$), and they are unable to exploit weight sharing~($8\times$), altogether $24\times$ energy saving is lost.

\section{Conclusion}

Fully-connected layers of deep neural networks
perform a matrix-vector multiplication.
For real-time networks where batching cannot be employed to improve re-use, these layers are
memory limited.
To improve the efficiency of these layers, one must reduce the energy needed to fetch their parameters.

This paper presents EIE, an energy-efficient engine optimized to operate on compressed deep neural networks. By leveraging sparsity in both the activations and the weights, and taking advantage of weight sharing and quantization, EIE reduces the energy needed to compute a typical FC layer by 3,400$\times$ compared with GPU. 
This energy saving comes from four main factors: the number of parameters is pruned by 10$\times$; Weight-sharing reduced the weights to only 4 bits. Then the smaller model can be fetched from SRAM and not DRAM, giving a 120$\times$ energy advantage; and since the activation vector is also sparse, only 30\% of the matrix columns need to be fetched for a final 3$\times$ savings. 
These savings enable an EIE PE to do 1.6 GOPS in an area of 0.64mm$^2$  and dissipate only 9mW. 64 PEs can process FC layers of AlexNet at 1.88$\times$10\textsuperscript{4} frames/sec.  
The architecture is scalable from one PE to over
256 PEs with nearly linear scaling of energy and
performance.
On 9 fully-connected layer benchmarks, EIE outperforms CPU, GPU and mobile GPU by factors of $189\times$, $13\times$ and $307\times$, and consumes $24,000\times$, $3,400\times$ and $2,700\times$ less energy than CPU, GPU and mobile GPU, respectively.

\newcommand{\BIBdecl}{\footnotesize\setlength{\itemsep}{0pt}}
\bstctlcite{bstctl:etal, bstctl:nodash, bstctl:simpurl}
\bibliographystyle{IEEEtran}
\bibliography{ref.bib}

\end{document}